\documentclass[runningheads]{llncs}
\usepackage{graphicx}
\usepackage{amsmath,amssymb} % define this before the line numbering.
\usepackage{color}
\usepackage{booktabs}
\usepackage{tabularx,ragged2e}
\usepackage{xcolor}
\usepackage[colorlinks = true]{hyperref}
\usepackage[width=122mm,left=12mm,paperwidth=146mm,height=193mm,top=12mm,paperheight=217mm]{geometry}
\begin{document}
% \linenumbers
\pagestyle{headings}
\mainmatter

\title{Modeling Context in Referring Expressions} % Replace with your title

\titlerunning{Modeling Context in Referring Expressions}

\authorrunning{Licheng et al.}

\author{Licheng Yu, Patrick Poirson, Shan Yang, Alexander C. Berg, Tamara L. Berg}

%Please write out author names in full in the paper, i.e. full given and family names. 
%If any authors have names that can be parsed into FirstName LastName in multiple ways, please include the correct parsing, in a comment to the volume editors:
%\index{Lastnames, Firstnames}
%(Do not uncomment it, because you may introduce extra index items if you do that...)

\institute{Department of Computer Science,\\
	University of North Carolina at Chapel Hill\\
	\email{ \{licheng,poirson,alexyang,aberg,tlberg\}@cs.unc.edu}
}

\maketitle

\begin{abstract}
Humans refer to objects in their environments all the time, especially in dialogue with other people.
We explore generating and comprehending natural language referring expressions for objects in images.
In particular, we focus on incorporating better measures of visual context into referring expression models and find that visual comparison to other objects within an image helps improve performance significantly.
We also develop methods to tie the language generation process together, so that we generate expressions for all objects of a particular category jointly.
Evaluation on three recent datasets - RefCOCO, RefCOCO+, and RefCOCOg\footnote{Datasets and toolbox can be downloaded from \href{https://github.com/lichengunc/refer}{https://github.com/lichengunc/refer}}, shows the advantages of our methods for both referring expression generation and comprehension.
\end{abstract}

\section{Introduction}

In this paper, we look at the dual-tasks of generating and comprehending natural language expressions referring to particular objects within an image.
Referring to objects is a natural and common experience.
For example, one often uses referring expressions in everyday speech to indicate a particular person or object to a co-observer, e.g., ``the man in the red hat'' or ``the book on the table''.
Computational models to generate and comprehend such expressions would have applicability to human-computer interactions,
%involving both visual and language cues, 
especially for agents such as robots, interacting with humans in the physical world.

Successful models will have to connect both recognition of visual attributes of objects and effective natural language generation to compose useful expressions for dialogue.  
A broader version of this latter goal was considered in 1975 by Paul Grice who introduced maxims describing cooperative conversation between people~\cite{Grice75}. 
These maxims, called the Gricean Maxims, describe a set of rational principles for natural language dialogue interactions. 
The 4 maxims are: quality (try to be truthful), quantity (make your contribution as informative as you can, giving as much information as is needed but no more), relevance (be relevant and pertinent to the discussion), and manner (be as clear, brief, and orderly as possible, avoiding obscurity and ambiguity).

\begin{figure}[t]
\centering
\includegraphics[width=0.68\textwidth]{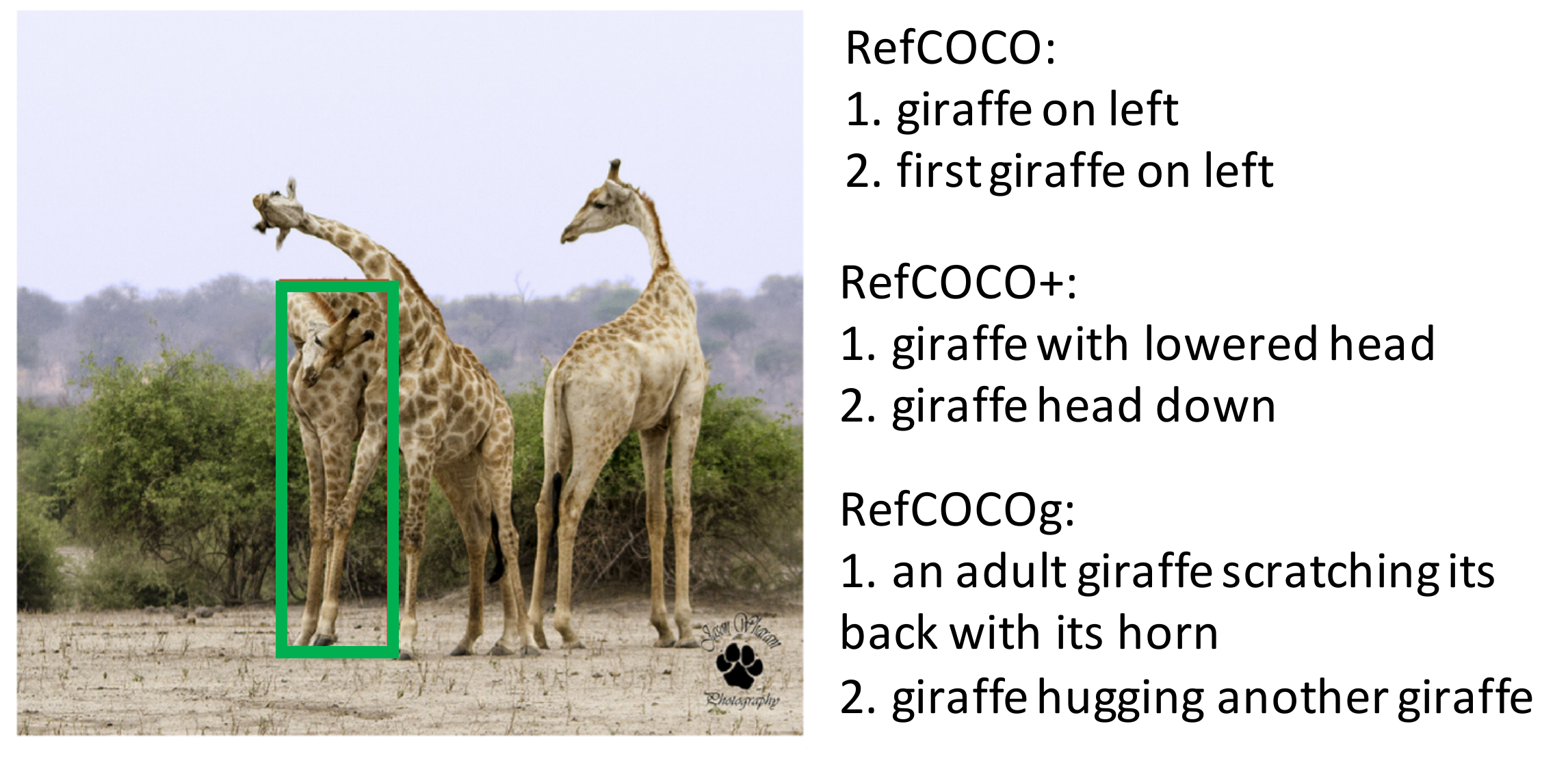}
\caption{Example referring expressions for the giraffe outlined in green from three referring expression datasets (described in Sec~\ref{sec:data}).}
\label{fig:giraffe}
\end{figure}

For the purpose of referring to objects in complex real world scenes these maxims suggest that a well formed expression should be informative, succinct, and unambiguous. 
The last point is especially necessary for referring to objects in the real world since we often find multiple objects of a particular category situated together in a scene. 
%Therefore, considering ambiguity during the generation process is highly important.
For example, consider the image in Fig.~\ref{fig:giraffe} which contains three giraffes. 
We should not refer to the target (outlined in green) as ``the spotted giraffe'' since all of the giraffes are spotted and this would create an ambiguous reference.
More reasonably we should refer to the target as ``the giraffe with lowered head'' to differentiate this giraffe from the other two.

The task of referring expression generation (REG) has been studied since the 1970s~\cite{winograd1972understanding,krahmer2012computational,mitchell2013generating,fitzgerald2013learning}, with most work focused on studying particular aspects of the problem in some relatively constrained datasets.
Recent approaches have pushed this work toword more realistic scenarios.
Kazemzadeh et al~\cite{kazemzadeh2014referitgame} introduced the first large-scale dataset of referring expressions for objects in real-world natural images, collected in a two-player game.
This dataset was originally collected on top of the 20,000 image ImageCleft dataset, but has recently been extended to images from the MSCOCO collection.
We make use of the RefCOCO and RefCOCO+ datasets in our work along with another recently collected referring expression dataset, released by Google, denoted in our paper as RefCOCOg~\cite{mao2015generation}.

The most relevant work to ours is Mao et al~\cite{mao2015generation} which introduced the first deep learning approach to REG.
In this model, the authors use a Convolutional Neural Network (CNN)~\cite{simonyan2014very} model pre-trained on ImageNet~\cite{russakovsky2015imagenet} to extract visual features from a bounding box around the target object and from the entire image.
They use these features plus 5 features encoding the target object location and size as input to a Long Short-term Memory (LSTM)~\cite{lstm} network that generates expressions.
Additionally, they apply the same model to the inverse problem of referring expression comprehension where the input is a natural language expression and the goal is to localize the referred object in the image.
% Another related work is from Hu et al~\cite{hu2015natural} which uses three LSTMs to encode the local appearance, global context and referring expressions jointly for the target object, but this model was only evaluated on referring expression comprehension task.

Similar to these recent methods, we also take a deep learning approach to referring expression generation and comprehension.
However, while they use a generic model for object context -- CNN features for the entire image containing the target object -- we take a more focused approach to encode object comparisons.
These object comparisons are critical for producing an unambiguous referring expression since one must consider visual characteristics of similar objects during generation in order to select the most distinct aspects for description.  
This mimics the process that a human would use to compose a good referring expression for an object, e.g. look at the object, look at other relevant objects, and generate an expression that could be used by a co-observer to unambiguously pick out the target object. 

In addition, for the referring expression generation task, we introduce a method to tie the language generation process together for all depicted objects of the same type.
This helps generate a good set of expressions such that the expressions differentiate between objects but are also complementary.
For example, we never want to generate the exact same expression for two objects in an image.
Alternatively, if we call one object ``the red ball'' then we may desire the expression for the other object to follow the same generation pattern, i.e., ``the blue ball''.  
Our experimental evaluations show that these visual and linguistic comparisons improve performance over previous state of the art.

In the rest of our paper, we first describe related work (Sec~\ref{sec:related}).  
We then describe our improvements to models for referring expression generation and comprehension (Sec~\ref{sec:model}), 
describe 3 referring expression datasets (Sec~\ref{sec:data}),
and perform experimental evaluations on several model variations (Sec~\ref{sec:experiments}).  
Finally we present our conclusions (Sec~\ref{sec:conclusion}).

\section{Related Work}
\label{sec:related}

Referring expressions are closely related to the more general problem of modeling the connection between images and descriptive language.
In recent years, this has been studied in the {\bf image captioning} task~\cite{farhadi2010every,socher2014grounded,hodosh2013framing,ordonez2011im2text,kulkarni2013babytalk}.
There, the aim is to condition the generation of language on the visual information from an image.
The wide range of aspects of an image that could be described, and the variety of words that could be chosen for a particular description complicate studying image captioning.
Our study of referring expressions is partially motivated by focusing on description for a specific, and more easily evaluated, communication goal.
Although our task is somewhat different, we borrow machinery from state of the art caption generation~\cite{donahue2015long,vinyals2015show,mao2014deep,fang2015captions,karpathy2015deep,kiros2014unifying,xu2015show} using LSTM to generate captions based on CNN features computed on an input image.
Three recent approaches for referring expression generation~\cite{mao2015generation} and comprehension~\cite{hu2015natural,rohrbach2015grounding} also take a deep learning approach.
However, we add visual object comparisons and tie together language generation for multiple objects.

{\bf Referring expression generation} has been studied for many years~\cite{winograd1972understanding,krahmer2012computational,mitchell2013generating} in linguistics and natural language processing.
These works were limited by data collection and insufficient computer vision algorithms.
Together Amazon Mechanical Turk and CNNs have somewhat mitigated these limitations, allowing us to revisit these ideas on large-scale datasets.
We still use such work to motivate the architecture of our pipeline.
For instance, Mitchell and Jordan et al~\cite{mitchell2013generating,jordan2000learning} show the importance of using attributes, Funakoshi et al~\cite{funakoshi2004generating} show the importance of relative relations between objects in the same perceptual group, and Kelleher et al~\cite{kelleher2006incremental} show the importance of spatial relationships.
These provide motivation for our modeling choices:
when considering a referring expression for an object, the model takes into account the relative spatial location of other objects of the same type and visual comparisons to objects in the same perceptual group.

{\bf The REG datasets} of the past were sometimes limited to using computer generated images~\cite{viethen2008use}, or relatively small collections of natural objects~\cite{mitchell2013typicality,mitchell2010natural,fitzgerald2013learning}.
Recently, a large-scale referring expression dataset was collected by Kazemzadeh et al~\cite{kazemzadeh2014referitgame} featuring natural objects in the real world.
Since then, another three REG datasets based on the object labels in MSCOCO have been collected~\cite{kazemzadeh2014referitgame,mao2015generation}.
The availability of large-scale referring expression datasets allows us to train deep learning models.
Additionally, our analysis of these datasets motivates our incorporation of visual comparisons between same-type objects, and the need to tie together choices for referring expression generation between objects.

\section{Models}
\label{sec:model}

We implement several model variations for referring expression generation and comprehension. 
The first set of models are recent state of the art deep learning approaches from Mao et al~\cite{mao2015generation}. 
We use these as our baselines (Sec~\ref{sec:baseline}). 
Next, we investigate incorporating better visual context features into the models (Sec~\ref{sec:visual}). 
Finally, we explore methods to jointly produce an entire set of referring expressions for all depicted objects of the same category (Sec~\ref{sec:language}). 

\begin{figure*}[t]
\centering
\includegraphics[width=0.68\textwidth]{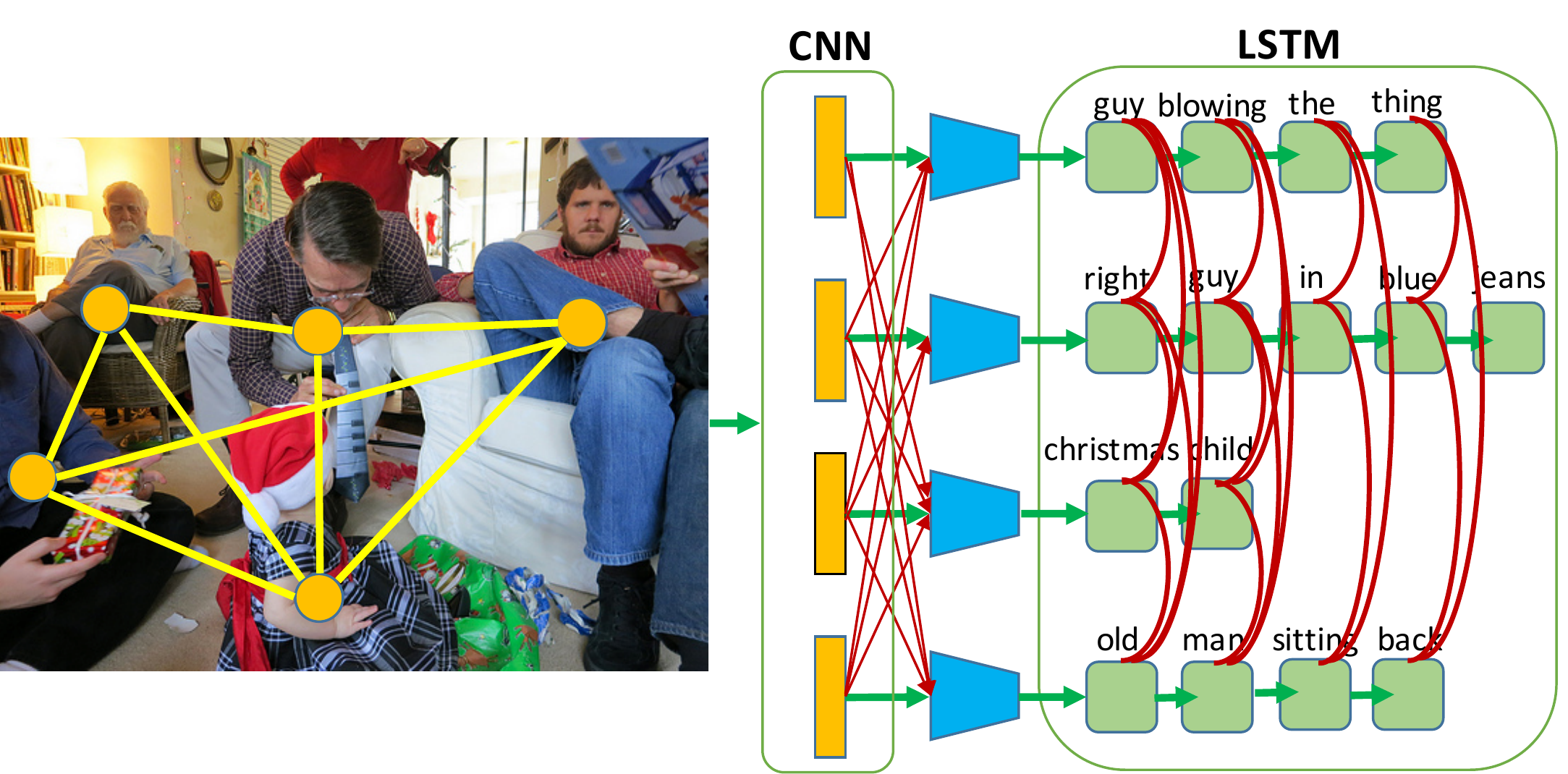}
\caption{Framework: We extract VGG-fc7 and location features for each object of the same type, then compute visual differences.
These features and differences are then fed into LSTM.
For sentence generation, the LSTMs are tied together, incorporating the hidden output difference as additional information for predicting words.
}
\label{fig:framework}
\end{figure*}

\subsection{Baselines}
\label{sec:baseline}
For comparison, we implement both the baseline and strong model of Mao et al~\cite{mao2015generation}.
Both models utilize a pre-trained CNN network to model the target object and its context within the image, and then use a LSTM for generation.
In particular, object and context are modeled as features from a CNN trained to recognize 1,000 object categories~\cite{simonyan2014very} from ImageNet~\cite{russakovsky2015imagenet}.
Specifically, the visual representation is composed of:
\begin{itemize}
\item Target object representation, $o_i$. 
The object is modeled as features extracted from the VGG-fc7 layer by forwarding its bounding box through the network.
\item Global context representation, $g_i$.
Context is modeled as features extracted from the VGG-fc7 layer for the entire image.
\item Location/size representation, $l_i$, for the target object.
Location and size are modeled as a 5-d vector encoding the x and y locations of the top left and bottom right corners of the target object bounding box, as well as the bounding box size with respect to the image, i.e., $l_i=[\frac{x_{tl}}{W}, \frac{y_{tl}}{H}, \frac{x_{br}}{W}, \frac{y_{br}}{H}, \frac{w\cdot h}{W\cdot H}]$.
\end{itemize}

Language generation is handled by a long short-term memory network (LSTM) \cite{lstm} where inputs are the above visual features and the network is trained to generate natural language referring expressions.
In Mao et al's baseline~\cite{mao2015generation}, the model uses maximum likelihood training and outputs the most likely referring expression given the target object, context, and location/size features.
In addition, they also propose a stronger model that uses maximum mutual information (MMI) training to consider whether a listener would interpret a referring expression unambiguously.
They impose this by penalizing the model if a generated referring expression could also be generated by some other object within the image.
We implement both their original model and MMI model in our experiments.
We subsequently refer to these two models as Baseline and MMI, respectively.

\subsection{Visual Comparison}
\label{sec:visual}
Previous works~\cite{brown2006watching,mitchell2013generating} have shown that objects in an image, of the same type as the target object, are most important for influencing what attributes people use to describe the target. 
One drawback of considering a general feature over the entire image to encode context (as in the baseline models) is that it may not specifically focus on visual comparisons to the most relevant objects -- the other objects of the same object category within the image. 

In this paper, we propose a more explicit encoding of the visual difference between objects of the same category within an image.
This helps for generating referring expressions which best discriminate the target object from the surrounding objects.
For example, in an image with three cars, two blue and one red, visual appearance comparisons could help generate ``the red car'' as an expression for the latter object.

Given the referred object and its surrounding objects, we compute two types of features for visual comparison.
The first type encodes the similarities and differences in {\em visual appearance} between the target object and other objects of the same cateogry depicted in the image.
Inspired by Sadeghi et al~\cite{sadeghi2015visalogy}, we compute the difference in visual CNN features as our representation of relative appearance.
Because there may be many surrounding objects of the same type in the image, and not every object will provide useful information about how to describe the target object, we need to first select which objects to compare and aggregate their visual differences.
In Section~\ref{sec:experiments}, we experiment with selecting different subsets of comparison objects:
objects of the same category, objects of different category, or all other depicted objects.
For each selected comparison object, we compute the appearance difference as the subtraction of the target object and comparison object CNN representations.
We experiment with three different strategies for computing an aggregate vector to represent the visual difference between the target object and the surrounding objects: minimum, maximum, and average over each feature dimension.
In our experiments, pooling the average difference between the target object and surrounding objects seems to work best.
Therefore, we use this pooling in all experiments.
\begin{itemize}
\item Visual appearance difference representation, $\delta v_i=\frac{1}{n}\sum_{j\neq i} \frac{o_i - o_j}{\| o_i - o_j \|}$, where $n$ is the number of objects chosen for comparisons and we use average pooling to aggregate the differences.
\end{itemize}

The second type of comparison feature encodes the {\em relative location and size} differences between 
the target object and surrounding objects of the same object category. 
People often use comparative size or location terms in referring expressions, e.g. ``the second giraffe from the left'' or ``the smaller monkey''~\cite{viethen2008use}. 
To address the dynamic number of nearby objects, we choose up to five comparison objects of the same category as the target object, sorted by distance to the target. 
When fewer than five objects of the same category are depicted, this 25-d vector (5-d x 5 surrounding objects) is padded with zeros. 
\begin{itemize}
\item Location difference representation, $\delta l_i$, where each 5-d difference is computed as 
% $\delta l_{ij}=[\frac{[\bigtriangleup x_{tl}]_{ij}}{W}, \frac{[\bigtriangleup y_{tl}]_{ij}}{H}, \frac{[\bigtriangleup x_{br}]_{ij}}{W}, \frac{[\bigtriangleup y_{br}]_{ij}}{H}, \frac{[\bigtriangleup A_{bbox}]_{ij}}{H\cdot W}]$.
$\delta l_{ij}=[\frac{[\bigtriangleup x_{tl}]_{ij}}{w_i}, \frac{[\bigtriangleup y_{tl}]_{ij}}{h_i}, \frac{[\bigtriangleup x_{br}]_{ij}}{w_i}, \frac{[\bigtriangleup y_{br}]_{ij}}{h_i}, \frac{w_j h_j}{w_i h_i}]$.
\end{itemize}

In summary, our final visual representation for a target object is:
\begin{equation}
r_i = W_m [o_i, g_i, l_i, \delta v_i, \delta l_i] + b_m
\end{equation}
where $o_i$, $g_i$, $l_i$ are the target object, global context, and location/size features from the baseline model, $\delta v_i$ and $\delta l_i$ encodes visual appearance difference and location difference.
$W_m$ and $b_m$ project the concatenation of the five types of features to be the final representation.

\subsection{Joint Language Generation}
\label{sec:language}

For the referring expression generation task, rather than generating sentences for each object in an image  separately~\cite{johnson2015densecap}\cite{mao2015generation}, we consider tying the generation process together into a single task to jointly generate expressions for all objects of the same object category depicted in an image. 
This makes sense intuitively -- when a person attempts to generate a referring expression for an object in an image they inherently compose that expression while keeping in mind expressions for the other objects in the picture. 
This can be observed in the fact that the expressions people generate for objects in an image tend to share similar patterns of expression. If you say ``the man on the left'' for one object then you tend to say ``the man on the right'' for the other object. We would like our algorithms to mimic these behaviors. 
Additionally, the algorithm should also be able to push generated expressions away from each other to create less ambiguous references. For example, if we use the word ``red'' to describe one object, then we probably shouldn't use the same word to describe another object. 
 
To model this joint generation process, we model generation using an LSTM 
model where in addition to the usual connections between time steps within an expression we also add connections between expressions for different objects. This architecture is illustrated in Fig~\ref{fig:framework}.

Specifically, we use LSTM to generate multiple referring expressions, $\left\{r_i\right\}$, given depicted objects of the same type, $\left\{o_j\right\}$.%, through $N$ time steps.
\begin{equation}
\begin{split}
P(R|O) &= \prod_i P(r_i|o_i, \{o_{j \neq i}\}, \{r_{j\neq i}\}),\\
&=
\prod_i\prod_tP(w_{i_t} | w_{i_{t-1}}, ..., w_{i_1}, v_i, \{h_{j_t, j\neq i}\})
\end{split}
\end{equation}
where $w_{i_t}$ are words at time $t$, $v_i$ visual representations, and $h_{j_t}$ is the hidden output of j-th object at time step t that encodes the visual and sentence information for the j-th object.
As visual comparison, we aggregate the difference of hidden outputs to push away ambiguous information.
$h_{dif_{i_t}} = \frac{1}{n}\sum_{j\neq i}\frac{h_{i_t}-h_{j_t}}{\|h_{i_t}-h_{j_t}\|}$.
There, $n$ is the the number of other objects of the same type.
The hidden difference is jointly embedded with the target object's hidden output, and forwarded to the softmax layer for predicting the word.
\begin{equation}
P(w_{i_t} | w_{i_{t-1}}, ..., w_{i_1}, v_i, \{h_{j_t, j\neq i}\}) = \mbox{softmax}(W_h [h_{i_t}, h_{dif_{i_t}}] + b_h)
\end{equation}

\section{Data}
\label{sec:data}

We make use of 3 referring expression datasets in our work, all collected on top of the Microsoft COCO image collection~\cite{lin2014microsoft}. One dataset, RefCOCOg~\cite{mao2015generation} is collected in a non-interactive setting, while the other two datasets, RefCOCO and RefCOCO+, are collected interactively in a two-player game~\cite{kazemzadeh2014referitgame}. 
In the following, we describe each dataset and provide some analysis of their similarities and differences, and then discuss splits of the datasets used in our experiments .

\subsection{Datasets \& Analysis}
Images for each dataset were selected to contain multiple objects of the same category (object categories depicted cover the 80 common objects from MSCOCO with ground-truth segmentation).
These images provide useful cases for referring expression generation since the referrer needs to compose a referring expression that uniquely singles out one object from other relevant objects.

\textbf{RefCOCOg:} This dataset was collected on Amazon Mechanical Turk in a non-interactive setting.
One set of workers were asked to write natural language referring expressions for objects in MSCOCO images then another set of workers were asked to click on the indicated object given a referring expression.
If the click overlapped with the correct object then the referring expression was considered valid and added to the dataset.
If not, another referring expression was collected for the object.
This dataset consists of 85,474 referring expressions for 54,822 objects in 26,711 images.
Images were selected to contain between 2 and 4 objects of the same object category.

\textbf{RefCOCO \& RefCOCO+:} These datasets were collected using the ReferitGame~\cite{kazemzadeh2014referitgame}.
In this two-player game, the first player is shown an image with a segmented target object and asked to write a natural language expression referring to the target object.
The second player is shown only the image and the referring expression and asked to click on the corresponding object.
If the players do their job correctly, they receive points and swap roles.
If not, they are presented with a new object and image for description.
Images in these collections were selected to contain two or more objects of the same object category.
In the RefCOCO dataset, no restrictions are placed on the type of language used in the referring expressions while in the RefCOCO+ dataset players are disallowed from using location words in their referring expressions by adding ``taboo'' words to the ReferItGame.
This dataset was collected to obtain a referring expression dataset focsed on purely appearance based description, e.g., ``the man in the yellow polka-dotted shirt'' rather than ``the second man from the left'', which tend to be more interesting from a computer vision based perspective and are independent of viewer perspective.
RefCOCO consists of 142,209 refer expressions for 50,000 objects in 19,994 images, and RefCOCO+ has 141,564 expressions for 49,856 objects in 19,992 images.

\textbf{Dataset Comparisons:}
As shown in Fig.~\ref{fig:giraffe}, the languages used in RefCOCO and RefCOCO+ datasets tend to be more concise and less flowery than the languages used in the RefCOCOg. 
RefCOCO expressions have an average length of 3.61 while RefCOCO+ have an average length of 3.53, and RefCOCOg contain an average of 8.43 words. 
This is most likely due to the differences in collection strategy. 
RefCOCO and RefCOCO+ were collected in a game scenario where players are trying to efficiently provide enough information to indicate the correct object to the other player. 
RefCOCOg was collected in independent rounds of Mechanical Turk without any interactive time constraints and therefore tend to provide more complex expressions, often entire sentences rather than phrases. 

In addition, RefCOCO and RefCOCO+ do not limit the number of objects of the same type to 4 and thus contain some images with many objects of the same type. 
Both RefCOCO and RefCOCO+ contain an average of 3.9 same-type objects per image, while RefCOCOg contains an average of 1.63 same-type objects per image. 
The large number of same-type objects per image in RefCOCO and RefCOCO+ suggests that incorporating visual comparisons to same-type objecs will be useful.

\textbf{Dataset Splits:}
There are two types of splits of the data into train/test sets: a per-object split and a people-vs-objects split.

The first type is \textbf{per-object split}.
In this split, the dataset is divided by randomly partitioning objects into training and testing sets.
This means that each object will only appear either in training or testing set, but that one object from an image may appear in the training set while another object from the same image may appear in the test set.
We use this split for RefCOCOg since same division was used in the previous state-of-the-art approach~\cite{mao2015generation}.

The second type is \textbf{people-vs-objects splits}.
One thing we observe from analyzing the datasets is that about half of the referred objects are people. 
Therefore, we create a split for RefCOCO and RefCOCO+ datasets that evaluates images containing multiple people (testA) vs images containing multiple instances of all other objects (testB). 
In this split all objects from an image will appear either in the training or testing sets, but not both. 
This split creates a more meaningfully separated division between training and testing, allowing us to evaluate the usefulness of context more fairly.

\section{Experiments}
\label{sec:experiments}

\begin{table}[b]
\scriptsize
\begin{center}
\begin{tabular}{|c|c|c|c|c|}
\hline
& \multicolumn{2}{c|}{RefCOCO} & \multicolumn{2}{c|}{RefCOCO+} \\
\cline{2-5}
& Test A & Test B & Test A & Test B \\
\hline
no context & 63.91\% & 66.31\% & 50.09\% & 45.05\% \\
global context & 63.15\% & 64.21\% & 48.73\% & 42.13\% \\
scale 2 & 65.57\% & 67.13\% & 50.38\% & 44.89\% \\
scale 3 & 66.14\% & 68.07\% & 50.25\% & 45.40\% \\
scale 4 & \bf{66.68\%} & \bf{68.56\%} & \bf{50.34\%} & \bf{45.48\%} \\
\hline
\end{tabular}
\end{center}
\caption{
Expression Comprehension accuracies on RefCOCO and RefCOCO+ of the Baseline model with differenct context source.
Scale $n$ indicates the size of the cropped window centered by the target object.
}\label{table:wo_context}
\end{table}

\begin{figure*}[b]
\centering
\includegraphics[width=0.8\linewidth]{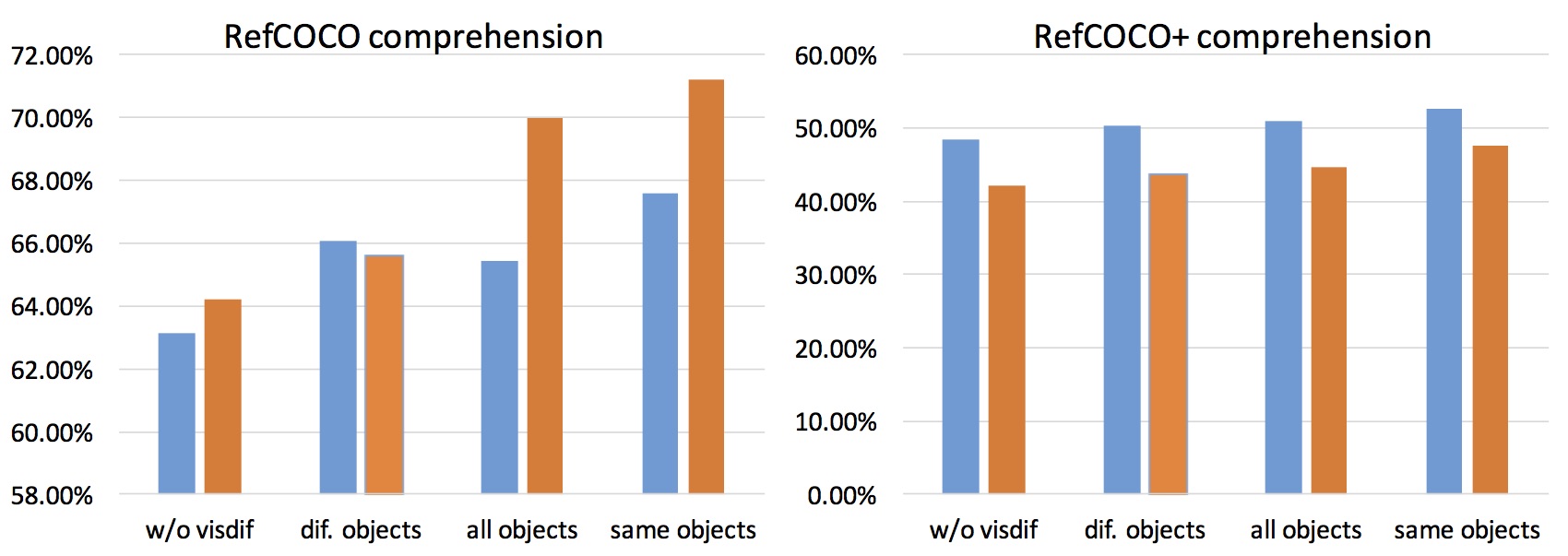}
\caption{Comprehension accuracies on RefCOCO and RefCOCO+ datasets. We compare the performance of ``visdif'' model without visual comparison, and visual comparison between different-category objects, between all objects, and between same-type objects.}\label{fig:tielangexamples}
\label{fig:different_context}
\end{figure*}

We first perform some experiments to analyze the use of context in referring expressions (Sec~\ref{sec:analysisexp}).
Given these findings, we then perform experiments evaluating the usefulness of our proposed visual and language innovations on the comprehension (Sec~\ref{sec:refcomprehension}) and generation tasks (Sec~\ref{sec:refgeneration}).

In experiments for the referring expression comprehension task, we use the same evaluation as Mao et al~\cite{mao2015generation}, namely we first predict the region referred by the given expression, then we compute the intersection over union (IOU) ratio between the true and predicted bounding box.
If the IOU is larger than 0.5 we count it as a true positive.
Otherwise, we count it as a false positive.
We average this score over all images.
For the referring expression generation task we use automatic evaluation metrics, BLEU, ROUGE, and METEOR developed for evaluating machine translation results, commonly used to evaluate language generation results~\cite{xu2015show,karpathy2015deep,fang2015captions,mao2014deep,vinyals2015show,kulkarni2013babytalk}.
We further perform human evaluations, and propose a new metric evaluating the duplicate rate of generated expressions.
For both tasks, we compare our models with ``Baseline'' and ``MMI''~\cite{mao2015generation}.
Specifically, we denote ``visdif'' as our visual comparison model, and ``tie'' as the LSTM tying model.
We also perform an ablation study, evaluating the combinations.

% BLEU measures truncated precision between the generated and ground truth referring expressions (1,2,3 indicate uni-gram, bi-gram, and tri-gram performance respectively). 
% ROUGE measures recall against the ground truth expression.
% METEOR measures the harmonic mean of precision and recall.

\subsection{Analysis Experiments}
\label{sec:analysisexp}

\subsubsection{Context Representation} 
As previously discussed, we suggest that the approaches proposed in recent referring expression works~\cite{mao2015generation,hu2015natural} make use of relatively weak contextual information, by only considering a single global image context for all objects.
To verify this intuition, we implemented both the baseline and strong MMI models from Mao et al~\cite{mao2015generation}, and compare the results for referring expression comprehension task with and without global context on RefCOCO and Refcoco+ in Table~\ref{table:wo_context}.
Surprisingly we find that the global context does not improve the performance of the model. 
In fact, adding context even decreases performance slightly. 
This may be due to the fact that the global context for each object in an image would be the same, introducing some ambiguity into the referring expression comprehension task.
Given these findings, we implemented a simple modification to the global context, computing the same visual representation, but on a somewhat scaled window centered around the target object. 
We found this to improve performance, suggesting room for improving the visual context feature. 
This motivate our development of a better context feature.
% Details about this experiment are reported in the supplementary file.

\subsubsection{Visual Comparison} 
For our visual comparison model, there could be several choices regarding which objects from the image should be compared to the target object.
We experiment with three sets of reference objects on RefCOCO and RefCOCO+ datasets: a) objects of the same-category in the image, b) objects of different-category in the image, and c) all objects appeared in the image.
We use our ``visdif'' model for this experiment.
The results are shown in Figure~\ref{fig:different_context}.
It is clear to see the visual comparisons to the same-category objects are most useful for referring expression comprehension task.
This is more like mimicing how human refer object -- we tend to point out the difference between the target object with the other same-category objects within the same image.

\subsection{Referring Expression Comprehension}
\label{sec:refcomprehension}

We evaluate performance on the referring expression comprehension task on RefCOCO, RefCOCO+ and RefCOCOg datasets.
For RefCOCO and RefCOCO+, we evaluate on the two subsets of people (testA) and all other objects (testB).
For RefCOCOg, we evaluate on the per-object split as previous work~\cite{mao2015generation}.
Since the authors haven't released their testing set, we show the performance on their validation set only, using the optimized hyper-parameters on RefCOCO.
Table~\ref{table:refcomprehension} shows the comprehension accuracies.
We observe that our implementation of Mao et al~\cite{mao2015generation} achieves comparable performance to the numbers reported in their paper. 
We also find that adding visual comparison features to the Baseline model improves performance across all datasets and splits.
Similar improvements are also observed on top of the MMI model.

In order to make a fully automatic referring system, we also train a Fast-RCNN~\cite{girshick2015fast} detector and build our system on top of the detections.
We train Fast-RCNN on the validation portion only as the RefCOCO and RefCOCO+ are collected using MSCOCO training data. For RefCOCOg, we use the detection results provided by~\cite{mao2015generation}, which were trained uisng Multibox~\cite{erhan2014scalable}.
Results on shown in the bottom half of Table~\ref{table:refcomprehension}.
Although all comprehension accuracies drop due to imperfect detections, the improvements of our models over Baseline and MMI are still observed.
One weakness of our automatic system is that it highly depends on detection performance, especially for general objects (testB).
However, considering our detector was trained on MSCOCO validation only, we believe such weakness may be alleviated with more training data and stronger detection techniques, e.g.,~\cite{he2015deep}\cite{liu2015ssd}\cite{ren2015faster}\cite{Zhu_DeePM_Arxiv15}\cite{bell2015inside}, etc.

\begin{figure*}
\centering
\includegraphics[width=1.0\linewidth]{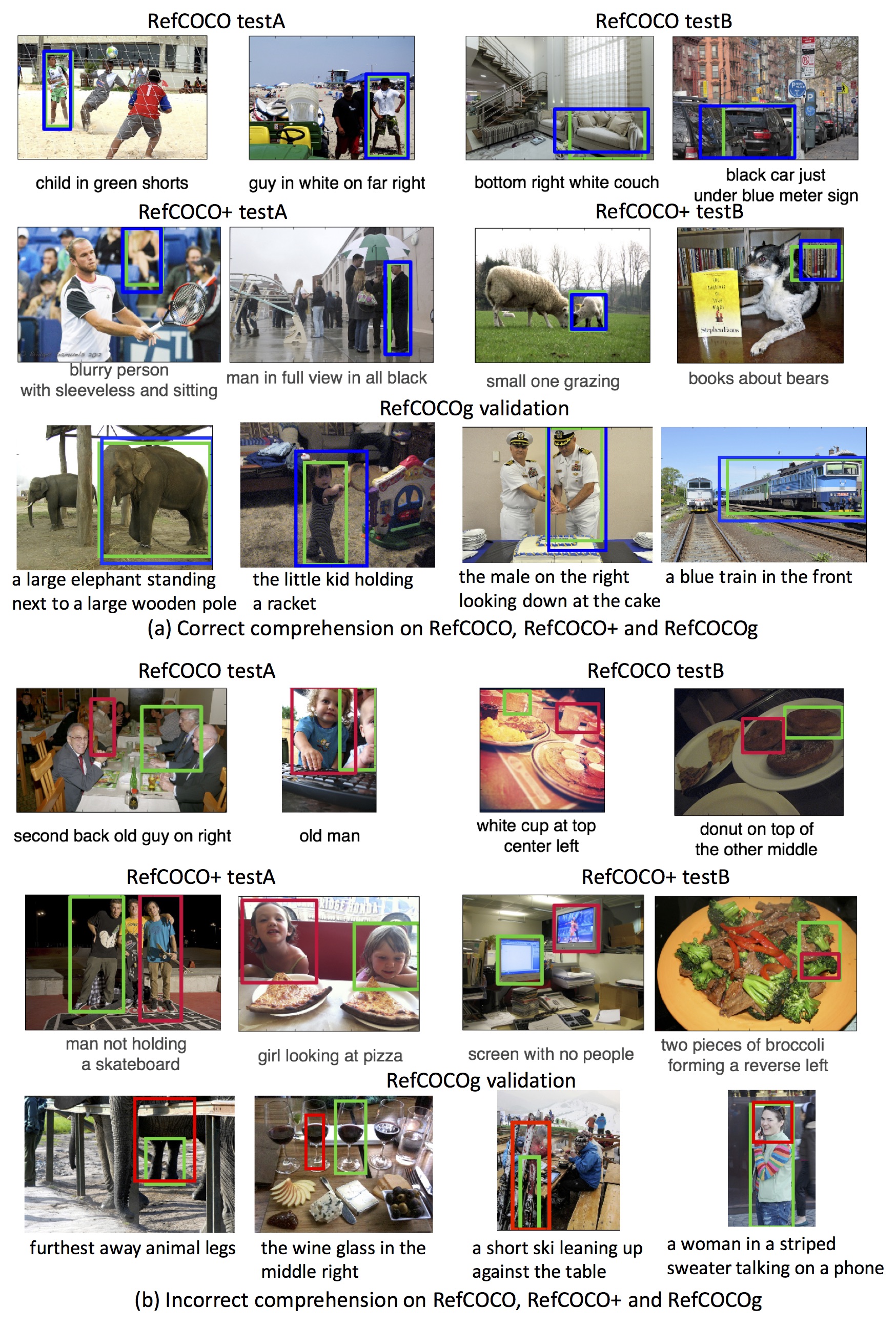}
\caption{Referring expression comprehension on RefCOCO and RefCOCO+ using ``visdif'' based on detections. The blue and red bounding boxes are correct and incorrect comprehension respectively, while the green boxes indicate the ground-truth regions.}\label{fig:auto_comprehension_refcoco_refcoco+_refcocog}
\end{figure*}

We show some automatic comprehension examples of RefCOCO, RefCOCO+ and RefCOCOg in Fig~\ref{fig:auto_comprehension_refcoco_refcoco+_refcocog}, where top three rows show correct comprehensions (object correctly localized) and bottom three rows show incorrect comprehensions (wrong object localized).

\begin{table*}[h]
\scriptsize
\begin{center}
\begin{tabular}{| l | c | c | c | c | c | c |}
\hline
&  \multicolumn{2}{c|}{RefCOCO} & \multicolumn{2}{c|}{RefCOCO+} & \multicolumn{1}{c|}{RefCOCOg}\\
\cline{2-6}
&\ \ Test A\ \  &\ \ Test B\ \   &\ \ Test A\ \  &\ Test B\ \ &\ \ Validation\ \  \\
\hline
Baseline\cite{mao2015generation}   		& 63.15\% 	& 64.21\%  	& 48.73\%  	& 42.13\% 	& 55.16\%\\
visdif   	& 67.57\%  	& 71.19\%   	& 52.44\%  	& 47.51\%	& 59.25\%\\
\hline
MMI\cite{mao2015generation} 		& 71.72\% 	& 71.09\% 	& 58.42\% 	& 51.23\%	& 62.14\%\\
visdif+MMI 	& \bf{73.98\%} 	& \bf{76.59\%} 	& \bf{59.17\%} 	& \bf{55.62\%}	& \bf{64.02\%} \\
\hline
\hline
Baseline(det)\cite{mao2015generation}   		& 58.32\% 	& 48.48\%  	& 46.86\%  	& 34.04\%   & 40.75\%\\
visdif(det)   									& 62.50\%  	& 50.80\%   & 50.10\%  	& 37.48\%	& 41.85\%\\
\hline
MMI(det)\cite{mao2015generation} 				& 64.90\% 	& 54.51\% 	& 54.03\% 	& 42.81\%	& 45.85\%\\
visdif+MMI(det) 								& \bf{67.64}\% 	& \bf{55.16}\% 	& \bf{55.81}\% 	& \bf{43.43}\%	& \bf{46.86}\%\\
\hline
\end{tabular}
\end{center}
\caption{Referring Expression comprehension results on the RefCOCO, RefCOCO+, and RefCOCOg datasets. Rows of ``method(det)'' are the results of automatic system built on Fast-RCNN~\cite{girshick2015fast} and Multibox~\cite{erhan2014scalable} detections.}
\label{table:refcomprehension}
\end{table*}

\subsection{Referring Expression Generation}
\label{sec:refgeneration}

\begin{figure*}
\centering
\includegraphics[width=1.0\linewidth]{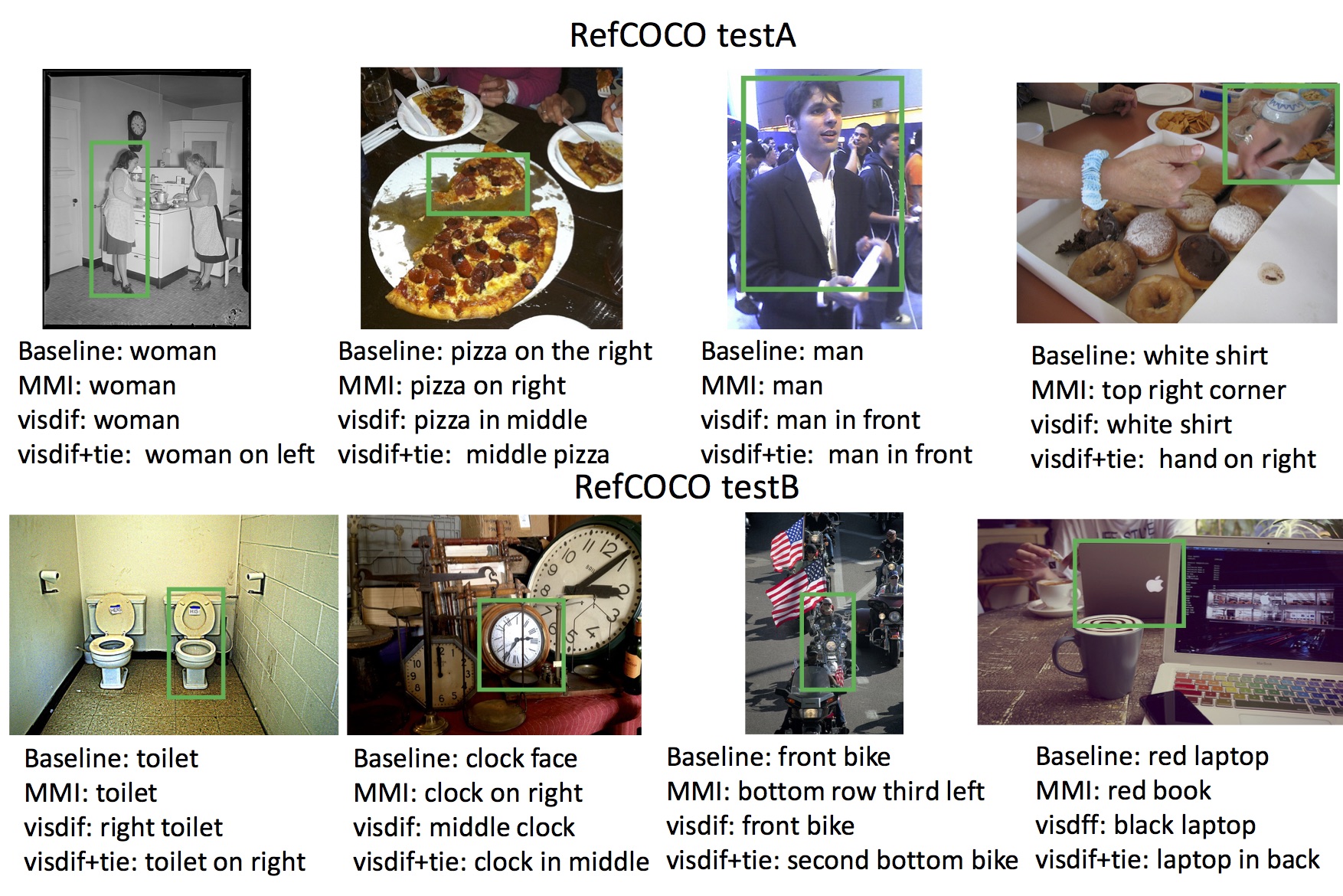}
\includegraphics[width=1.0\linewidth]{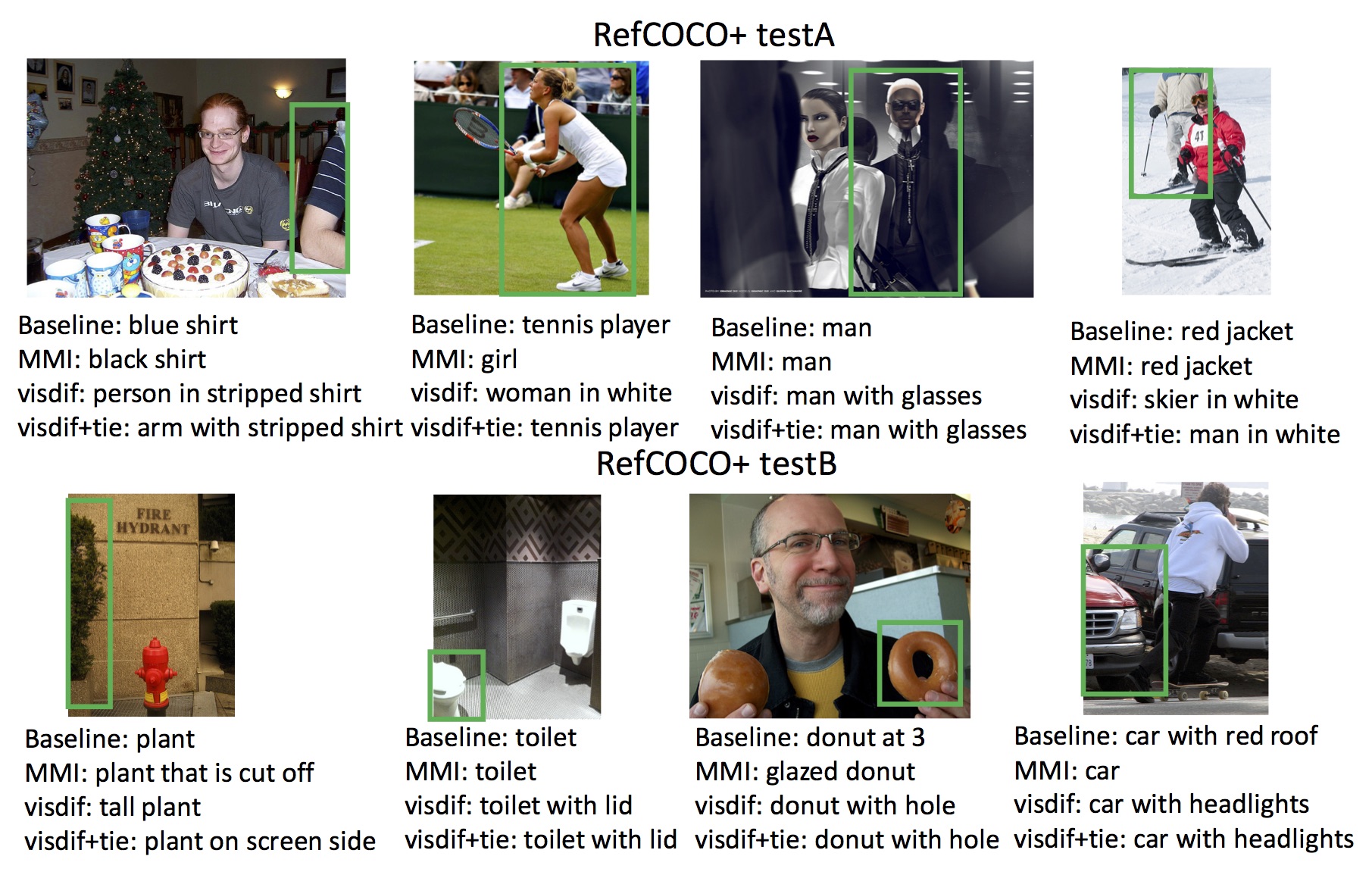}
\caption{Referring expression generation on RefCOCO and RefCOCO+ by different methods.}\label{fig:generation}
\end{figure*}

For the referring expression generation task, we evaluate the usefulness of our visual comparison features as well as our joint language generation model.
These serve to tie the generation process together so that the model considers other objects of the same type both visually and linguistically during generation.
On the visual side, comparisons are used to judge similarity of the target object to other objects of the same type in terms of appearance, size and location.
On the language side, the joint LSTM model serves to both differentiate and mimic language patterns in the referring expressions for the entire set of depicted objects.
Fig~\ref{fig:generation} shows some comparison between our model with other methods.

Our full results are shown in Table~\ref{table:generation}. 
We find that incorporating our visual comparison features into the Baseline model improves generation quality (compare row ``Baseline'' to row ``visdif'').
It also improves the performance of MMI model (compare row ``MMI" to row ``visdif+MMI'').
We also observe that tying the language generation together across all objects consistently improves the performance (compare the bottom three ``+tie'' rows with the above).
Especially for method ``visdif+tie'', it achieves the highest score under almost every measurement.
We do not perform language tying on RefCOCOg since here some objects from an image may appear in training while others may appear in testing.

\vspace{-0.5cm}
\begin{table*}
\scriptsize
\centering
\begin{tabular}{| c | c | c | c | c | c | c | c | c |}
\multicolumn{9}{c}{RefCOCO} \\
\hline
&  \multicolumn{4}{c|}{Test A} & \multicolumn{4}{c|}{Test B}\\
\cline{2-9}
&\ \ Bleu 1\ \  &\ \ Bleu 2\ \   &\ \ Rouge\ \  &\ Meteor\ \ &\ \ Bleu 1\ \  &\ \ Bleu 2\ \   &\ \ Rouge\ \  &\ Meteor\ \ \\
\hline
Baseline~\cite{mao2015generation}& 0.477 & 0.290 & 0.413 & 0.173 & 0.553 & 0.343 & 0.499 & 0.228 \\
MMI~\cite{mao2015generation}  	 & 0.478 & 0.295 & 0.418 & 0.175 & 0.547 & 0.341 & 0.497 & 0.228 \\
\hline
visdif 							 & 0.505 & \bf{0.322} & 0.441 & 0.184 & 0.583 & 0.382 & 0.530 & 0.245 \\
visdif+MMI 						 & 0.494 & 0.307 & 0.441 & 0.185 & 0.578 & 0.375 & 0.531 & 0.247 \\
\hline
Baseline+tie 					 & 0.490 & 0.308 & 0.431 & 0.181 & 0.561 & 0.352 & 0.505 & 0.234 \\
visdif+tie 						 & \bf{0.510} & 0.318 & \bf{0.446} & \bf{0.189} & \bf{0.593} & \bf{0.386} & \bf{0.533} & \bf{0.249} \\
visdif+MMI+tie  				 & 0.506 & 0.312 & 0.445 & 0.188 & 0.579 & 0.370 & 0.525 & 0.246 \\
\hline
\end{tabular}

\begin{tabular}{| c | c | c | c | c | c | c | c | c |}
\multicolumn{9}{c}{RefCOCO+} \\
\hline
&  \multicolumn{4}{c|}{Test A} & \multicolumn{4}{c|}{Test B}\\
\cline{2-9}
&\ \ Bleu 1\ \  &\ \ Bleu 2\ \   &\ \ Rouge\ \  &\ Meteor\ \ &\ \ Bleu 1\ \  &\ \ Bleu 2\ \   &\ \ Rouge\ \  &\ Meteor\ \ \\
\hline
Baseline~\cite{mao2015generation}& 0.391 & 0.218 & 0.356 & 0.140 & 0.331 & 0.174 & 0.322 & 0.135 \\
MMI~\cite{mao2015generation}  	 & 0.370 & 0.203 & 0.346 & 0.136 & 0.324 & 0.167 & 0.320 & 0.133 \\
\hline
visdif 							 & 0.407 & 0.235 & 0.363 & 0.145 & 0.339 & 0.177 & 0.325 & 0.145 \\
visdif+MMI 						 & 0.386 & 0.221 & 0.360 & 0.142 & 0.327 & 0.172 & 0.325 & 0.135 \\
\hline
Baseline+tie 					 & 0.392 & 0.219 & 0.361 & 0.143 & 0.336 & 0.177 & 0.325 & 0.140 \\
visdif+tie 						 & \bf{0.409} & \bf{0.232} & \bf{0.372} & \bf{0.150} & \bf{0.340} & \bf{0.178} & \bf{0.328} & \bf{0.143} \\
visdif+MMI+tie  				 & 0.393 & 0.220 & 0.360 & 0.142 & 0.327 & 0.175 & 0.321 & 0.137 \\
\hline
\end{tabular}

\begin{tabular}{| c | c | c | c | c |}
\multicolumn{5}{c}{RefCOCOg} \\ 
\hline
& \multicolumn{4}{c|}{validation}\\
\cline{2-5}
& Bleu 1 & Bleu 2 & Rouge & Meteor \\
\hline
Baseline~\cite{mao2015generation} & 0.437  & 0.273  & 0.363  & 0.149 \\
MMI~\cite{mao2015generation} 	  & 0.428  & 0.263  & 0.354  & 0.144 \\
\hline
visdif 							  & \bf{0.442}  & \bf{0.277}  & \bf{0.370}  & \bf{0.151} \\
visdif+MMI 						  & 0.430  & 0.262  & 0.356  & 0.145 \\
\hline
\end{tabular}

\caption{Referring Expression Generation Results: Bleu, Rouge, Meteor evaluations for RefCOCO, RefCOCO+ and RefCOCOg. }
\label{table:generation}
\end{table*}
\vspace{-0.5cm}

\begin{figure*}
\centering
\includegraphics[width=1.0\linewidth]{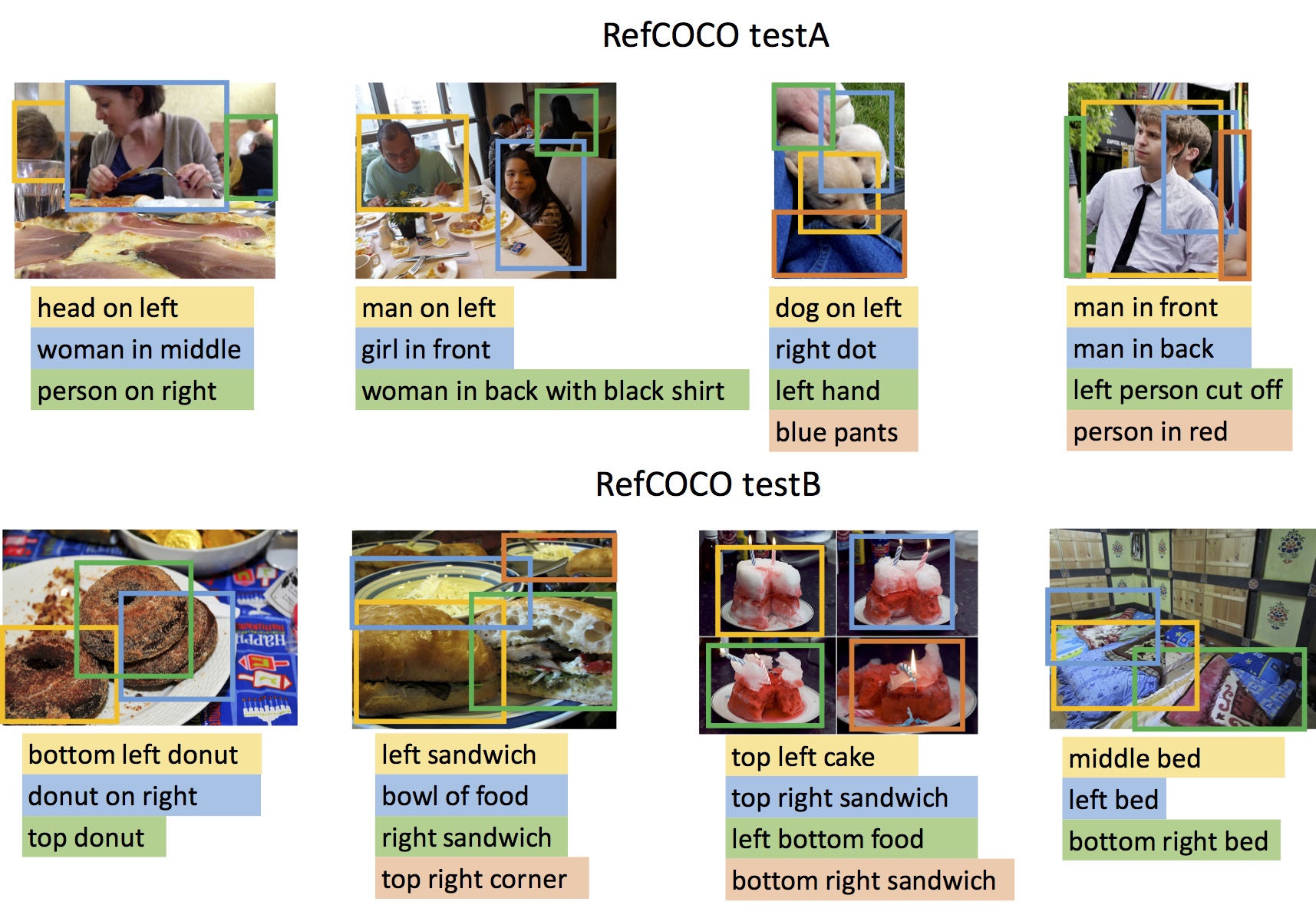}
\includegraphics[width=1.0\linewidth]{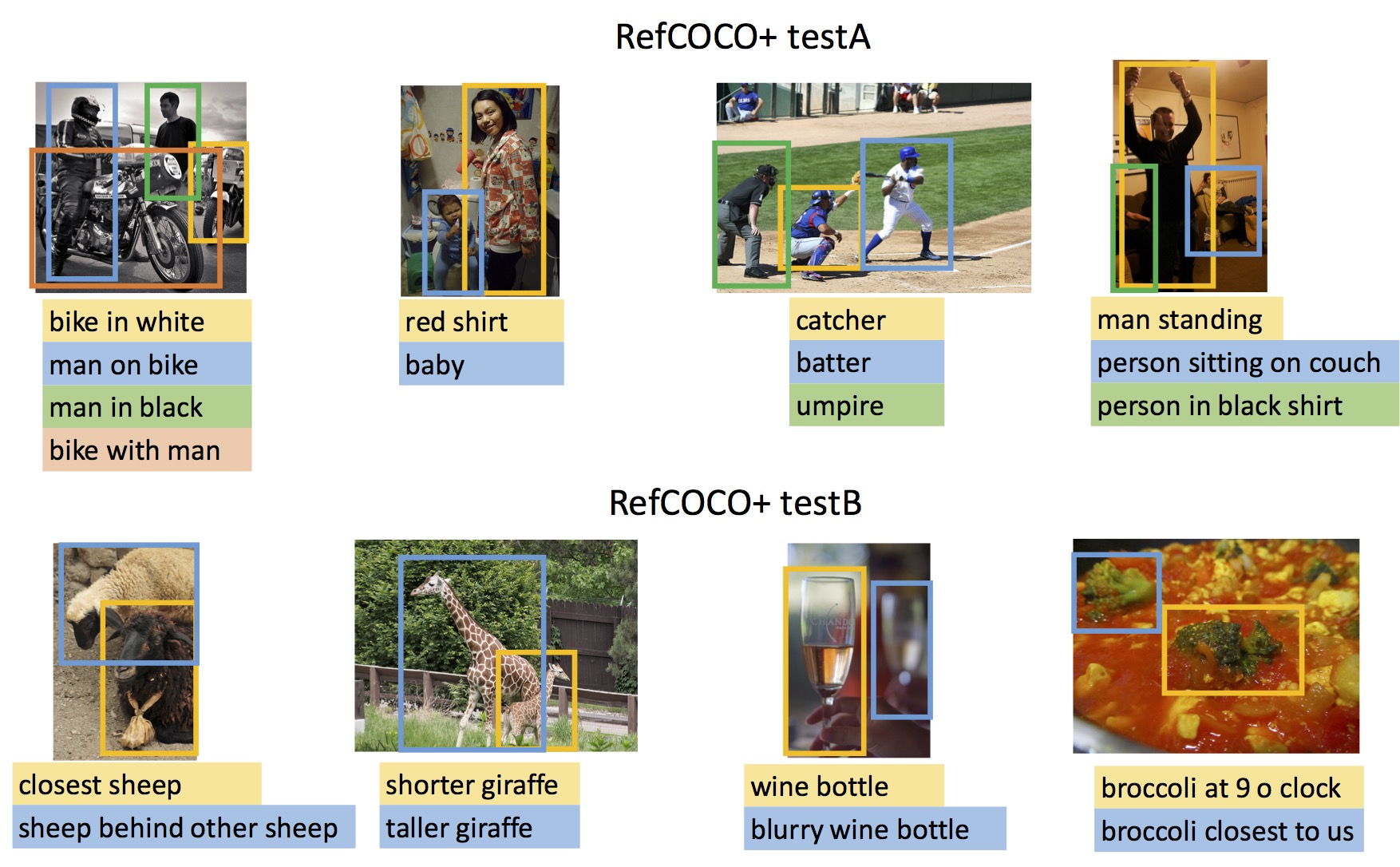}
\caption{Joint referring expression generation using our full model of ``visdif+MMI+tie''.}\label{fig:tielang}
\end{figure*}

We observe in Table~\ref{table:generation} that models incoporating ``+MMI'' are worse than without ``+MMI'' under the automatic scoring metrics.
To verify whether these metrics really reflect performance, we performed human evaluations on the expression generation task.
Three Turkers were asked to click on the referred object given the image and the generated expression.
If more than two clicked on the true target object, we consider this expression to be correct.
Table~\ref{table:human_evaluation} shows the human evaluation results, indicating that models with ``+MMI'' are consistently higher performance.
%than without considering listener's comprehension.
We also find ``+tie'' methods perform the best, indicating that tying language together is able to produce less ambiguous referring expressions.
% Fig~\ref{fig:tielangexamples} shows examples of tied generations. %demonstrates our method for tying language together.
Some referring expression generation examples using different methods are shown in Fig~\ref{fig:generation}.
% Besides, in Fig~\ref{fig:tielang}, we show more examples of ``visdif+MMI+tie'' which generates sentences for all referred objects within an image together using tied language model.
Besides, we show more examples of tied generations using ``visdif+MMI+tie'' model in Fig~\ref{fig:tielang}.

Finally, we introduce another evaluation metric which measures the fraction of images for which an algorithm produces the same generated referring expression for multiple objects within the image.
Obviously, a good referring expression generator should never produce the same expressions for two objects within the same image. 
Thus we would like this number to be as small as possible.
The evaluation results under such metric are shown in Table~\ref{table:duplicate}.
We find ``+MMI'' produces smaller number of duplicated expressions on both RefCOCO and RefCOCO+, while ``+tie'' helps generating even more different expressions.
Our combined model ``visdif+MMI+tie'' performs the best under this metric. 

\begin{table*}[t]
\scriptsize
\centering
\begin{tabular}{| c | c | c | c | c |}
\hline
& \multicolumn{2}{c}{RefCOCO} & \multicolumn{2}{c|}{RefCOCO+} \\
\cline{2-5}
&\ \ Test A\ \  &\ \ Test B\ \   &\ \ Test A\ \  &\ Test B\ \\
\hline
Baseline~\cite{mao2015generation} & 62.42\% & 64.99\% & 49.18\% & 42.03\% \\
MMI 							  & 65.76\% & 68.25\% & 49.84\% & 45.38\% \\
\hline
visdif 							  & 68.27\% & 74.92\% & 55.20\% & 43.65\% \\
visdif+MMI 						  & 70.25\% & 75.47\% & 53.56\% & 47.58\% \\
\hline
Baseline+tie 					  & 64.51\% & 68.34\% & 52.06\% & 43.53\% \\
visdif+tie 						  & \bf{71.40\%} & 76.14\% & \bf{57.17\%} & 47.92\% \\
visdif+MMI+tie 					  & 70.01\% & \bf{76.31\%} & 55.64\% & \bf{48.04}\% \\
\hline
\end{tabular}
\caption{Human Evaluations on referring expression generation.}
\label{table:human_evaluation}
\end{table*}

\begin{table*}[t]
\scriptsize
\centering
\begin{tabular}{| c | c | c | c | c |}
\hline
& \multicolumn{2}{c}{RefCOCO} & \multicolumn{2}{c|}{RefCOCO+} \\
\cline{2-5}
&\ \ Test A\ \  &\ \ Test B\ \   &\ \ Test A\ \  &\ Test B\ \\
\hline
Baseline~\cite{mao2015generation} & 15.60\% & 16.40\% & 28.67\% & 46.27\% \\
MMI 							  & 11.60\% & 11.73\% & 21.07\% & 26.40\% \\
\hline
visdif 							  &  9.20\% &  8.80\% & 19.60\% & 31.07\% \\
visdif+MMI 						  &  5.07\% &  6.13\% & 12.13\% & 16.00\% \\
\hline
Baseline+tie 					  & 11.20\% & 14.93\% & 22.00\% & 32.13\% \\
visdif+tie 						  &  \bf{4.27\%} &  5.33\% & 11.73\% & 16.27\% \\
visdif+MMI+tie 					  &  6.53\% &  \bf{4.53\%} & \bf{10.13\%} & \bf{13.33\%} \\
\hline
\end{tabular}
\caption{Fraction of images for which the algorithm generates the same referring expression for multiple objects. Smaller is better.}
\label{table:duplicate}
\end{table*}

\vspace{-0.4cm}
\section{Conclusion}
\label{sec:conclusion}
\vspace{-0.4cm}
In this paper, we have developed a new model for incorporating detailed context into referring expression models. With this visual comparison based context we have improved performance over previous state of the art for referring expression generation and comprehension. In addition, for the referring expression generation task, we explore methods for joint generation over all relevant objects. Experiments verify that this joint generation improves results over previous attempts to reduce ambiguity during generation. 

\noindent
{\bf Acknowledgements:}
We thank Junhua Mao, Dequan Wang and Varun K. Nagaraja for helpful discussions. 
This research is supported by NSF Grants \#1444234, 1445409, 1405822, and Microsoft.

\bibliographystyle{splncs03}
\bibliography{egbib}
\end{document}